\documentclass[11pt]{article}

\usepackage[preprint]{acl}
\usepackage{adjustbox}
\usepackage{times}
\usepackage{latexsym}
\usepackage{amsfonts}
\usepackage{amssymb}
\usepackage{multirow}
\usepackage{graphics}
\usepackage{tabularx}
\usepackage{arydshln}
\usepackage{array} 
\usepackage{url}
\usepackage{graphicx}
\usepackage{wrapfig}
\usepackage{tabularx}
\usepackage[T1]{fontenc}

\usepackage[utf8]{inputenc}

\usepackage{microtype}

\usepackage{inconsolata}

\title{Team QUST at SemEval-2025 Task 10: Evaluating Large Language Models in Multiclass Multi-label Classification of News Entity Framing}

\author{Jiyan Liu\textsuperscript{1}, Youzheng Liu\textsuperscript{1}, Taihang Wang\textsuperscript{1}, Xiaoman Xu\textsuperscript{1}\thanks{Corresponding author}, Yimin Wang\textsuperscript{2} and Ye Jiang\textsuperscript{1}  \\
        College of Information Science and Technology\textsuperscript{1} \\ 
        College of Data Science\textsuperscript{2} \\
        Qingdao University of Science and Technology, China}
        
\begin{document}
\maketitle
\begin{abstract}
This paper introduces the participation of the QUST team in subtask 1 of SemEval-2025 Task 10. We evaluate various large language models (LLMs) based on instruction tuning (IT) on subtask 1. Specifically, we first analyze the data statistics, suggesting that the imbalance of label distribution made it difficult for LLMs to be fine-tuned. Subsequently, a voting mechanism is utilized on the predictions of the  top-3 models to derive the final submission results. The team participated in all language tracks, achieving 1st place in Hindi (HI), 2nd in Russian (RU), 3rd in Portuguese (PT), 6th in Bulgarian (BG), and 7th in English (EN) on the official test set. We release our system code at: \url{https://github.com/warmth27/SemEval2025_Task10}
\end{abstract}

\section{Introduction}
SemEval-2025 Task 10 encourages participants to develop algorithms for multilingual recognition and extraction of narrative tasks from online news \cite{semeval2025task10, guidelinesSE25T10}. We participated in subtask 1 (Entity Framing), which aims to assign fine-grained role labels, covering three main types (protagonists, antagonists, and innocent) to named entities (NEs) mentioned in news articles, based on a predefined fine-grained role classification system \cite{mahmoud2025entity}. This is a multi-label multi-class text-span classification task. 

During this process, we face several challenges:
\begin{itemize}
    \item \textbf{Complexity of languages: }Compared to widely spoken languages like English, smaller languages such as Bulgarian lack high-quality pre-trained language models, which significantly heighten the complexity of model architecture design. 
    \item \textbf{Data scarcity: }Insufficient training data has resulted in suboptimal fine-tuning outcomes for the large language models (LLMs). Performance is also weaker for languages with smaller datasets.
\end{itemize}

In subtask 1, we first conduct a quick evaluation by comparing our model with the baseline model. Inspired by \citet{xu2024team} and \citet{zhang2023instruction}, we apply instruction tuning (IT) to several LLMs on subtask 1. However, IT entails substantial training costs. We first fine-tuning the models in English, then apply it to other languages to minimize fine-tuning and evaluation time. 

To enhance the model's performance and generalization ability, we adopt a hard voting based on ensemble learning \cite{jabbar2024advanced}, in which the top-3 selected models vote to determine the final prediction. Experimental results indicate that ensemble learning significantly enhances the model performance. 

\section{Related Work}
\subsection{Instruction tuning for LLMs}
IT is a powerful technique that adjusts the input context to align with specific instructions, updating the parameters of LLMs in a supervised manner \cite{wang2024instruction,jiang2023team}. Current studies frequently examine the efficacy of IT for LLMs. For instance, \citet{qin2024unleashing} presents a comprehensive review of IT in LLMs, detailing the fine-tuning process with instruction pairs and evaluating the critical factors that influence the outcomes of IT. 
\citet{wang2024survey} emphasizes that in the IT process of LLMs, the quality of the dataset plays a more significant role than its quantity.

Similar to the aforementioned studies, in subtask 1, IT helps the model understand complex contextual information and accurately assign fine-grained roles to entities based on instructions. By applying customized IT, we can improve the model’s performance on complex tasks, particularly when large-scale labeled data is scarce, as this method effectively enhances the model's generalization ability. 

\subsection{Voting} 
Voting is an ensemble learning strategy that enhances overall performance by combining the predictions of several base models \cite{xu2024team,abro2021vote}. 
The voting strategy we adopt is hard voting, in which the final result is determined by the majority vote based on the predicted class labels of the top-3 models. The most frequent class is chosen as the final outcome. 

\begin{table}[h]
\setlength{\tabcolsep}{13pt} 
\renewcommand{\arraystretch}{1.2}
\centering
\begin{tabular}{l|ccc}
\hline
\textbf{Language} & \textbf{train} & \textbf{dev} & \textbf{test} \\ \hline
BG       & 627   & 31  & 124  \\ \hline
EN       & 686   & 91  & 235  \\ \hline
HI       & 2331  & 280 & 316  \\ \hline
PT       & 1251  & 116 & 297  \\ \hline
RU       & 722   & 86  & 214  \\ \hline
\end{tabular}
\caption{Statistics of each language}
\label{table 1}
\end{table}

\section{Experimental Setup}
\subsection{Data}

\begin{figure}[b]
\centering
  \includegraphics[width=\columnwidth]{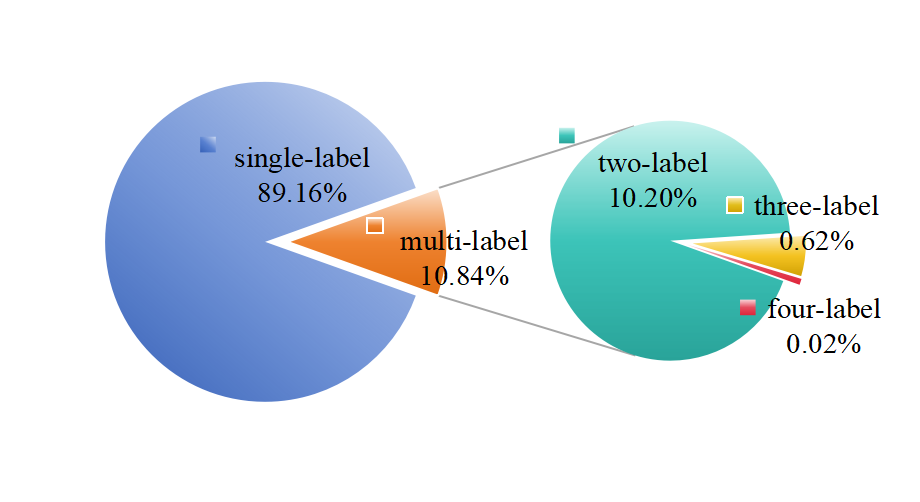}
  \caption{Distribution of the number of labels in the dataset.}
  \label{fig:1}
\end{figure}

The statistical distribution of each language dataset is presented in Table \ref{table 1}. It is evident that the data distribution is imbalanced across the five languages, especially for languages with fewer training samples (such as BG and EN), which could impact the model's generalization ability.

Additionally, we analyze the distribution of label counts, as shown in Figure \ref{fig:1}. The results indicate that the majority of samples are single-labeled. Based on this observation, we design an experiment in which the task is treated as a single-label task to evaluate whether this approach can enhance the original results. The experimental results and analysis are presented in Section \ref{single}.

\begin{figure*}[t]
\centering
  \includegraphics[width=\textwidth]{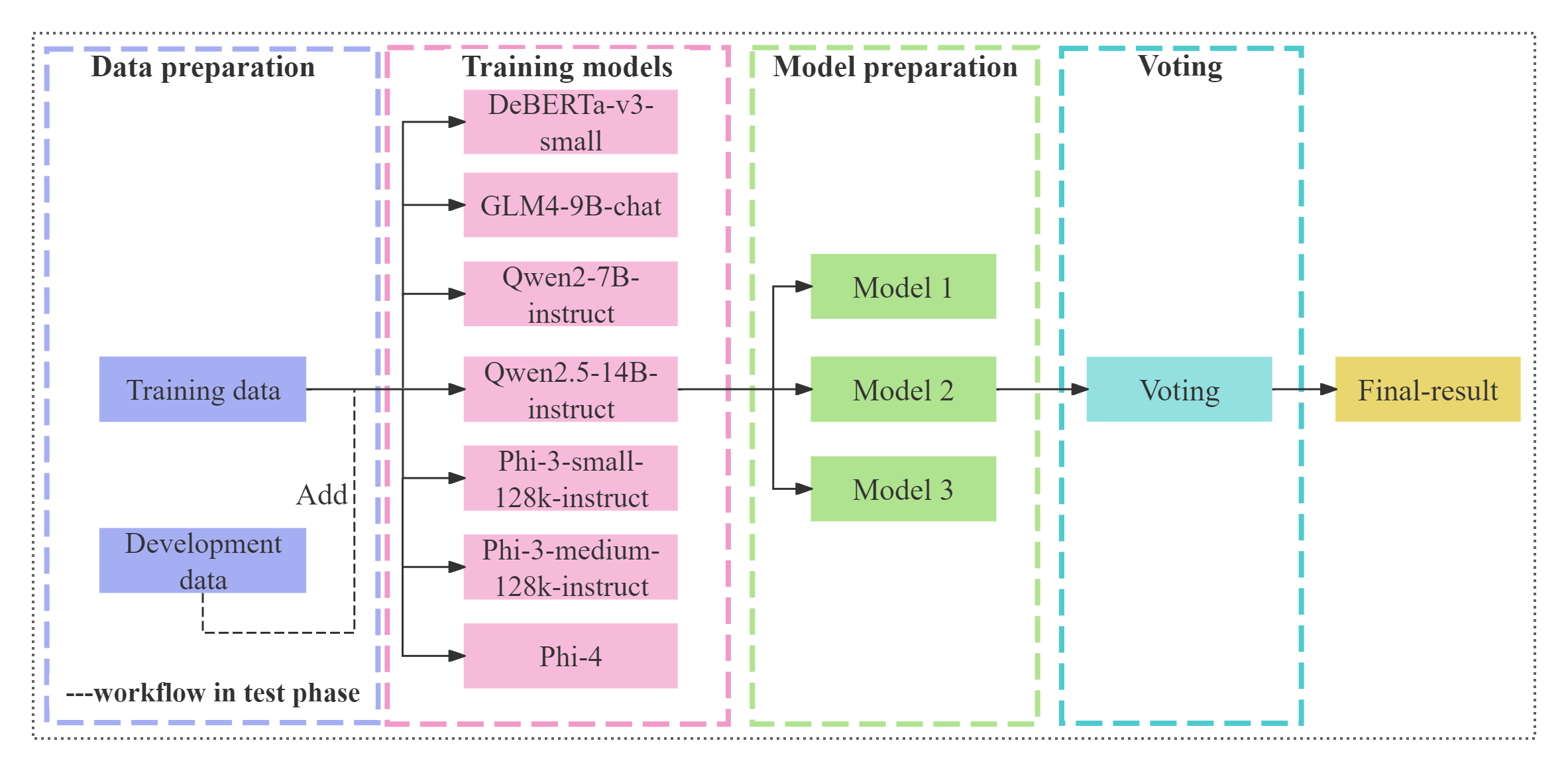}
  \caption{llustration of the overall workflow in this paper. "Model 1", "Model 2" and "Model 3" represent the top-3 best models of the performance employed.}
  \label{fig:2}
\end{figure*}

\subsection{Instruction strategy}
Inspired by \citet{wang2024instruction}, we improve several versions based on their work. Ultimately, we find that this version of the instruction yields the best results. The example directive is: "\textit{Given an article and an entity within that article. Analyze this article and the entity, and provide the fine-grained roles of the entity.}" This instruction emphasizes the requirement to assignone or more fine-grained roles to each entity, supporting a one-to-many label output. 

\begin{table*}[t]
\centering
\begin{tabular}{c|c|cc}
\hline
\textbf{Language}            & \textbf{Model}                                      & \textbf{Methods}      & \textbf{Exact Match Ratio} \\ \hline
\multirow{2}{*}{EN} & \multirow{2}{*}{Phi-3-small-128k-instruct} & Single-label & 0.3736            \\
                    &                                            & Multi-label  & \textbf{0.3846 }           \\ \hline
\end{tabular}
\caption{The result of Single-label Vs. Multi-label. \textbf{Bold} indicates the best result.}
\label{Table 2}
\end{table*}

\begin{table*}[h]
\centering
\renewcommand{\arraystretch}{1.2}
\begin{tabular}{l|lc}
\hline
\textbf{Methods}                              & \textbf{Models}                        & \textbf{Exact Match Ratio} \\ \hline
Small language model                 & DeBERTa-v3-small               & 0.2747            \\ \hline
\multirow{6}{*}{Instruction tuning} & GLM4-9B-chat                   & 0.1978            \\
                                     & Qwen2-7B-instruct             & 0.3626            \\
                                     & Qwen2.5-14B-instruct          & 0.3626            \\
                                     & Phi-3-small-128k-instruct     & 0.4505            \\
                                     & Phi-3-medium-128k-instruct    & 0.4505            \\
                                     & Phi-4                         & \underline{0.4615}        \\ \hline
Voting                               & Phi-3-small+Phi-3-medium+Phi4 & \textbf{0.4725 }           \\ \hline
\end{tabular}
\caption{Evaluation of methods and models on English development data. \textbf{Bold} indicates the best result, and \underline{Underline} indicates the second-best result.}
\label{Table 3}
\end{table*}

\subsection{Model configurations}
Before identifying the final models, We first compared the performance of several baseline models from Hugging Face\footnote{https://huggingface.co/}. Considering the multilingual nature of the task, we primarily conduct experiments using the English dataset during the model comparison phase. The models involved in the evaluation include DeBERTa-v3-small \cite{he2021debertav3}, GLM4-9B-chat \cite{glm2024chatglm}, Qwen2-7B-instruct \cite{yang2024qwen2}, Qwen2.5-14B-instruct \cite{yang2024qwen2}, Phi-3-small-128k-instruct (Phi-3-small) \cite{abdin2024phi}, Phi-3-medium-128k-instruct (Phi-3-medium) \cite{abdin2024phi}, and Phi-4 \cite{abdin2024Phi-4}. 

Given the variations in data size across languages, as well as considerations for training time and computational efficiency, we established two training schemes: 10 epochs and 20 epochs. The learning rate for the Qwen2-7B-instruct model is set to 1e-5, while for the other models, it is set to 1e-4. To avoid overfitting and enhance storage efficiency, we implement an epoch selection strategy, retaining only the model parameters that yield the best performance. 

The final result utilizes an ensemble learning strategy, employing hard voting to combine the predictions of the top-3 selected models for each language, thereby enhancing the prediction accuracy of the final result.

\section{Results}

\subsection{Single-label result}
\label{single}
Since the majority of samples are single-labeled, we extract data containing only single labels from the English training dataset to treat this task as a single label task. As shown in Table \ref{Table 2}, the score for the single-label task is 0.3736, whereas the score for the multi-label task is 0.3846. The result for the single-label task drops by only 2.86\%. Although this approach does not improve model performance, the decline is minimal, suggesting a significant imbalance in the dataset.

\subsection{Evaluation results}
Our evaluation primarily relies on the official development data. The goal of our evaluation is to rapidly identify models and methods that perform well, and to compare the performance variations among different models. To enhance experimental efficiency, we focus primarily on evaluating the English dataset at this stage.

As shown in Table \ref{Table 3}, all large language models except GLM4-9B-chat significantly outperform DeBERTa-v3-small, confirming that IT can effectively enhance the performance of LLMs on this task. Among these, GLM4-9B-chat yields the lowest result, possibly due to limited instruction comprehension, which leads to weaker generalization ability.

Phi-4 achieves the best performance among the single models, slightly outperforming Phi-3-medium and Phi-3-small, suggesting that larger-scale Phi-series models offer more stability on this task.

Voting strategy further enhances the final performance by 1.1\% compared to the Phi-4, indicating that the voting strategy effectively integrates the advantages of the individual models. However, the improvement is limited, possibly due to similar predictions across models or the inherent performance ceiling of the task.

\subsection{Official test results}
\begin{table}[]
\centering
\setlength{\tabcolsep}{13pt}   
\renewcommand{\arraystretch}{1.2}
\begin{tabular}{l|cc}
\hline
\multicolumn{1}{c|}{\multirow{2}{*}{\textbf{Language}}} & \textbf{Baseline}  & \textbf{QUST(rank)}   \\
\multicolumn{1}{c|}{}                          & subtask 1 & subtask 1    \\ \hline
BG                                             & 0. 0403   & 0. 3871(6th) \\ \hline
EN                                             & 0. 0383   & 0. 3277(7th) \\ \hline
HI                                             & 0. 0570   & \textbf{0. 4684(1st)} \\ \hline
PT                                             & 0. 0471   & 0. 4579(3rd) \\ \hline
RU                                             & 0. 0514   & \underline{0. 5140(2nd)} \\ \hline
\end{tabular}
\caption{Official test results.\textbf{Bold} indicates the best result, and \underline{Underline} indicates the second-best result.}
\label{Table 4}
\end{table}

Our approach participated in several language tracks, with Hindi ranked 1st, Russian ranked 2nd, Portuguese ranked 3rd, and Bulgarian and English ranked 6th and 7th, respectively, as shown in Table \ref{Table 4}. Additionally, our approach significantly outperforms the baseline of subtask 1 across all languages, demonstrating the effectiveness of our approach. However, it is worth noting that our performance in English and Bulgarian is comparatively weaker, while the final test results for Hindi are better than the performance in the previous evaluation phase.

This unexpected phenomenon may be related to the scale of the training data. The training data for English and Bulgarian is relatively limited, which may limit the model's generalization ability, while Hindi benefits from a richer dataset, allowing the model to better learn task patterns and perform more effectively in the test phase. Furthermore, Table \ref{Table 4} shows strong performance in Russian and Portuguese, further indicating that the scale of training data might be a key factor affecting model performance.

\section{Conclusion}
In summary, our team developed an effective approach for subtask 1 of SemEval-2025 Task 10. We first conduct instruction tuning on large language models on the English dataset and choose the models that perform best, then adapt them to other languages. In the final testing phase, to augment the training data, we incorporate the development data into the training set and select the top-performing model based on evaluation. Ultimately, hard voting successfully integrates the advantages of several models, thereby enhancing the prediction accuracy.

As the training data for English and Bulgarian is relatively limited and we have not yet employed data augmentation methods, future work will explore effective strategies to augment both the quantity and quality of data, thereby enhancing the model's capacity to comprehend texts from diverse languages and cultural backgrounds. 

\section*{Acknowledgements}
This work is funded by the Natural Science Foundation of Shandong Province under grant ZR2023QF151 and the Natural Science Foundation of China under grant 12303103.


\bibliography{acl}





\end{document}